\documentclass[runningheads,a4paper]{llncs}

\usepackage{amssymb}
\usepackage{graphicx}

\usepackage{url}
\urldef{\mailsa}\path|{a.alansary14}@imperial.ac.uk|

\usepackage{amsmath}
\usepackage{booktabs}
\usepackage{subfig}
\usepackage{color}
\usepackage{hyperref}

\begin{document}

\mainmatter  % start of an individual contribution

\title{Automatic View Planning with Multi-scale Deep Reinforcement Learning Agents}
\titlerunning{Automatic View Planning with Multi-scale Deep RL-Agents}


\author{Amir Alansary%
% \thanks{Corresponding author.}%
\and Loic Le Folgoc \and Ghislain Vaillant \and Ozan Oktay \and \\ 
Yuanwei Li \and Wenjia Bai \and Jonathan Passerat-Palmbach \and Ricardo Guerrero \and \\
Konstantinos Kamnitsas \and Benjamin Hou \and Steven McDonagh \and Ben Glocker \and Bernhard Kainz \and Daniel Rueckert}
% \author{**** \\ ****}
%
\authorrunning{A. Alansary et al.}
% \authorrunning{PaperID 1066}

\institute{Imperial College London,UK \\
\mailsa
}
% \institute{PaperID 1066}

\toctitle{Lecture Notes in Computer Science}
\tocauthor{Authors' Instructions}
\maketitle

% =============================================================================
% =============================================================================

\begin{abstract}
We propose a fully automatic method to find standardized view planes in 3D image acquisitions. Standard view images are important in clinical practice as they provide a means to perform biometric measurements from similar anatomical regions. These views are often constrained to the native orientation of a 3D image acquisition. Navigating through target anatomy to find the required view plane is tedious and operator-dependent. For this task, we employ a multi-scale reinforcement learning (RL) agent framework and extensively evaluate several Deep $Q$-Network (DQN) based strategies. RL enables a natural learning paradigm by interaction with the environment, which can be used to mimic experienced operators. We evaluate our results using the distance between the anatomical landmarks and detected planes, and the angles between their normal vector and target. The proposed algorithm is assessed on the mid-sagittal and anterior-posterior commissure planes of brain MRI, and the 4-chamber long-axis plane commonly used in cardiac MRI, achieving accuracy of $1.53$mm, $1.98$mm and $4.84$mm, respectively.
% \keywords{View Planning, Reinforcement Learning, Deep Learning, Deep Q-Network, DQN, Cardiac, Brain}
\end{abstract}
% =============================================================================
% =============================================================================
% \vspace{-0.6cm}
\section{Introduction}
In medical imaging, obtaining accurate biometric measurements that are comparable across populations is essential for diagnosis and supporting critical decision making. For this purpose, standard view planes through a defined anatomy are commonly used in clinical practice to establish comparable metrics. Finding these planes in an imaging examination through a 3D volume is slow and suffers from inter-observer variability.
The neuro-imaging community defines a standard (axial) image plane by adopting the anterior-posterior commissure (ACPC) line. Transforming an image to the ACPC coordinate system includes a number of steps: (i) marking the AC point, (ii) obtaining the optimal view of the ACPC and the mid-sagittal plane, and (iii) marking the PC point. Accurate detection of the mid-sagittal plane is useful for the initial step in image registration~\cite{ardekani1997automatic}. It is also used in evaluation of pathological brains by estimating the departures from bilateral symmetry in the cerebrum~\cite{stegmann2005mid}. Similarly, in cardiac MRI standard views are used to assess anomalies. Because of the complexity of cardiac anatomy, the appearance of relevant structures can exhibit large variance according to the positioning of the imaging plane. During conventional cardiac MRI acquisition, the localization of short and long-axis of the heart requires a multi-step approach that involves double-oblique slices, exhibiting both inter and intra-observer variance~\cite{lu2011automatic}. These steps include: (i) whole 3D pilot image acquisition, (ii) left ventricle (LV) localization, (iii) short axis orientation, (iv) 3-chamber view calculation, (v) landmark detection in mid-ventricular slices, and (vi) 4- and 2-chamber view calculation.

In this work, we aim to automate the view planning process by using reinforcement learning (RL) where an agent learns to make comprehensive and sensible decisions by mimicking navigation processes as outlined above, in a manner that allows medical experts to gain confidence in fully automatic methods. %RL enables learning to solve complex problems by dividing the solution into sequential steps, where it is challenging to learn from an external supervision.
RL constitutes a sub-field of machine learning concerned with how agents take actions in an environment. In contrast to supervised learning, RL involves learning by interacting with an environment instead of using a set of labeled examples that is typically provided by a knowledgeable supervisor. This learning paradigm allows RL agents to learn complex tasks that may need several steps to find a solution~\cite{sutton1998reinforcement}.
% Early work~\cite{sahba2006reinforcement} proposed an RL approach for medical image segmentation that found optimal thresholds using a reinforcement algorithm.
% Learning an optimal RL policy essentially results in learning a mapping from each current state to an action that maximizes the expected rewards. 
% The advent of deep learning has fuelled the currently highly active RL research field. 
Mnih et al.~\cite{mnih2015human} adopted a deep convolution network for RL function approximation, known as the Deep $Q$-Network (DQN), achieving human-level performance in a suite of Atari games. Recently, DQN has shown promising results when employed in related applications in the medical imaging domain. Ghesu et al.~\cite{ghesu2017multi} introduced an automatic landmark detection approach using a DQN-agent to navigate in 3D images with fixed step actions. Maicas et al.~\cite{maicas2017deep} proposed a similar method for breast lesion detection using actions to control the location and size of the bounding box. Liao et al.~\cite{liao2017artificial} presented an image registration approach using actions to explore transformation parameters. We adopt different DQN-based architectures as a solution for the proposed RL formulation of the view planning task. 

% \textbf{Automatic View Planning:}
\textbf{Related Work:}
% \textbf{Adult Brain MRI:}
Ardekani et al.~\cite{ardekani1997automatic} proposed a method to automatically detect the mid-sagittal plane in 3D brain images by maximizing the cross-correlation between the two image sections on either side of the sought plane. Stegmann et al. \cite{stegmann2005mid} proposed to use a sparse set of profiles in the plane normal direction and maximize the local symmetry around them.
% , instead of finding global symmetry~\cite{ardekani1997automatic}.
In~\cite{le2017computationally,lu2011automatic}, they proposed an automatic view planning algorithm for cardiac MRI acquisition. Their methods are based on learning the anatomy segmentation and detecting anchor landmarks in order to calculate standard cardiac views. These methods require prior knowledge of the whole 3D image for the purpose of plane detection.
% They also rely on detecting landmarks, which may complicate the pipeline and propagate errors to latter stages. 
This involoves manual annotation of anatomical landmarks, which is a tedious and time-consuming task. In our method, we use the acquired standardized views for cardiac scans in training without any manual labeling.

\textbf{Contribution:}
%\todoJon{I think it's worth itw own subsection}
We propose a novel RL-based approach for fully automatic standard view plane detection from volumetric MRI data. The proposed model follows a multi-scale search strategy with hierarchical action steps in a coarse-to-fine fashion. By sequentially updating plane parameters, our algorithm is able to reach the target plane. We run extensive experiments for evaluating different DQN baselines on detecting 3 different planes. Applications of our method to brain and cardiac MRI data show a target plane detection in real time with accuracy around $2$ and $5$ mm, respectively. 
%Apply our algorithm to (***) and show results on (***) exhibiting (***). 
% =============================================================================
% =============================================================================
\section{Background}
An RL agent learns by interacting with an environment, $E$. %In the canonical framework, 
At every state, $s$, a single decision is made to choose an action, $a$, from a set of multiple discrete actions, $A$. Each valid action choice results in an associated scalar reward, defining the reward signal, $R$. The agent attempts to learn a policy to maximize both immediate and subsequent future rewards (optimal policy).

\textbf{$Q$-Learning:}
The optimal action-selection policy can be identified by learning a state-action value function, $Q(s,a)$~\cite{watkins1992q}. The $Q$-function is defined as the expected value of the accumulated discounted future rewards $E [ r_{t+1} + \gamma r_{t+2} + \dots + \gamma r_{t+n} | s,a ]$. $\gamma \in [0,1]$ is a discount factor that represents the uncertainty in the agent's environment and is used to weight future rewards accordingly. This value function can be unrolled recursively (using the Bellman Equation~\cite{bellman2013dynamic}) and can thus be solved iteratively: $Q_{i+1}(s,a) = E \left[ r + \gamma \: \underset{a'}{\max}\: Q_{i}(s',a')\right]$.

% BH edit
%The optimal action-selection policy can be identified by learning the $Q$-function, $Q(s,a) = E \left[ r_{t+1}+ \gamma r_{t+2} + \gamma^2 r_{t+3} + \dots | s,a \right]$, where $a$ and $s$ are the action and state respectively, and $\gamma \in [0,1]$ is a discount factor that represents the uncertainty in the agent's environment. Thus it is being used to weight future rewards accordingly. This value function can be unrolled recursively (using the Bellman Equation~\cite{bellman2013dynamic}) and can thus be solved iteratively: $Q_{i+1}(s,a) = E \left[ r + \gamma \: \underset{a'}{\max}\: Q_{i}(s',a')\right]$.

% =============================================================================

\textbf{Deep $Q$-Learning: }
Mnih et al.~\cite{mnih2015human} proposed the Deep $Q$-Network (DQN) and implemented a standard $Q$-learning algorithm with the addition of approximating the $Q$-function using a ConvNet, $Q(s,a) \approx Q(s,a;\omega)$, where $\omega$ represents the network's parameters. The DQN loss function is defined as:
\[
L(\omega) = E \left[ \left( r + \gamma \: \underset{a'}{\max}\: Q_{target}(s',a';{\omega }^{-}) - Q_{net}(s,a;\omega) \right)^{2} \right],
\]
\noindent Approximating the $Q$-function in this manner allows to learn from larger data sets using mini-batches. The DQN uses $Q_{target}(\omega^{-})$, a fixed version of $Q_{net}(\omega)$, that is periodically updated. This is used to stabilize rapid policy changes, due to the quick variations in $Q$-values and the distribution of the data. Another problem that may cause divergence is successive data sampling. To avoid this, an experience replay memory %~\cite{lin1993reinforcement}
that stores transitions of $(s_{t},a_{t},r_{t+1},s_{t+1})$ is randomly sampled to create the mini-batches used for training. We outline below two recent state-of-the-art improvements to the standard DQN.%, and evaluate them experimentally in Section~\ref{sec_experiments}. 

\textbf{Double DQN (DDQN):} It has been shown that DQN is susceptible to bias in noisy environments, where the target network may cause upward bias due to delayed updates. Van Hasselt et al.~\cite{van2016deep} proposed a solution that replaces the maximum approximated action from $Q_{target}(s',a';{\omega }^{-})$ with an action selected from the $Q_{target}\left(s',Q_{net}(s',a',\omega);{\omega }^{-}\right)$. This strategy is able to mitigate bias by decoupling the selected action from $Q_{target}$. DDQN improves the stability of learning, which can translate to the ability to learn more complicated tasks but may not necessarily improve the performance~\cite{van2016deep}.

\textbf{Duel DQN:} Wang et al.~\cite{wang2015dueling} showed improved performance over the original DQN by defining two separate channels: \textit{(i)} an action-independent value function $V(s)$ to provide an estimate of the value of each state, and \textit{(ii)} an action-dependent advantage function $A(s,a)$ to calculate potential benefits of each action. Both functions are then combined into a single action-advantage $Q$-function, $Q(s,a)=A(s,a)+V(s)$. Duel DQN may achieve more robust estimates of state value by decoupling it from specific actions, so $s$ could be more explicitly modelled, which yields higher performance in general. 

% =============================================================================
% =============================================================================
\section{Method}
% \subsection{RL Formulation for Automatic View Planning}
\begin{figure}[!htb]
    % \vspace{-1cm}
    \centering
    \includegraphics[width=0.90\textwidth]{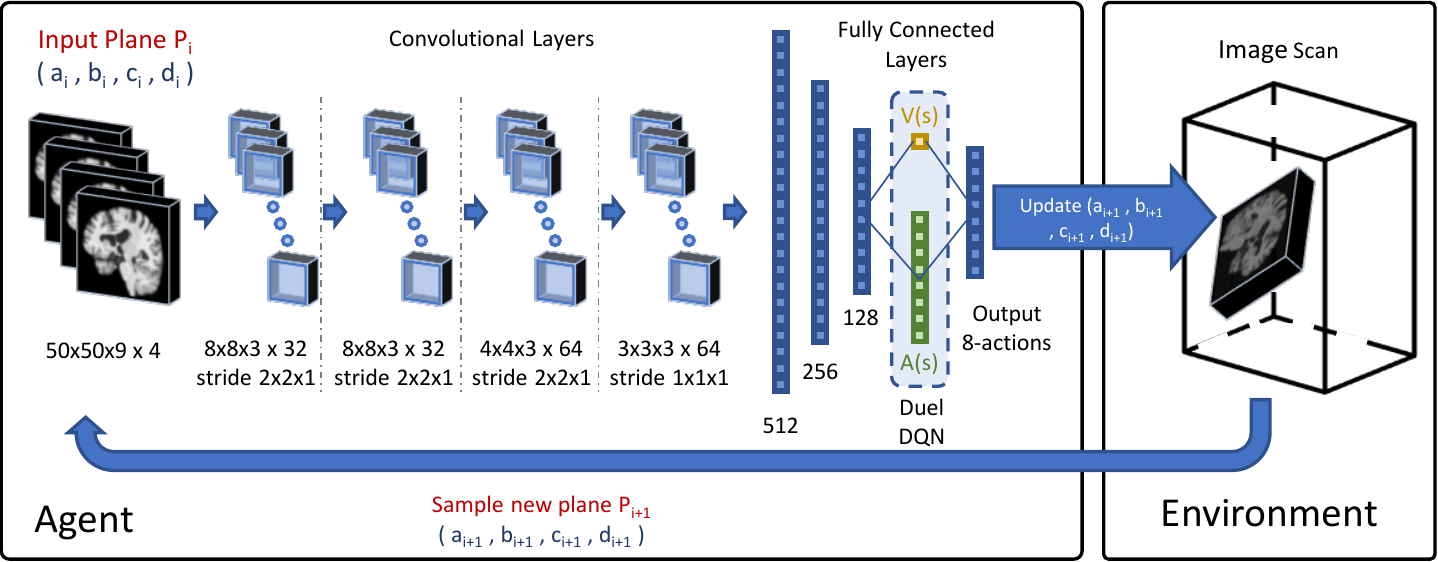}
    \caption{The pipeline of the proposed multi-scale RL agent. Initially, the environment samples a plane: $ax + by + cz + d = 0$, from the 3D image scan. The agent selects an action to update a single parameter for sampling the next plane. This process is repeated till the agent reaches a terminal state (oscillation).}
    \label{fig_framework}
    % \vspace{-0.75cm}
\end{figure}
A plane $P$, in the Cartesian coordinate frame of the 3D image, is defined as: $ax + by + cz + d = 0$. Where $(a,b,c)$ represent the normal direction (cosine) to this plane and $d$ is the distance of the plane from the origin. To automate standard view planning, we aim to find the appropriate parameterization of the target plane. We formulate our RL framework by defining the following elements:

\begin{itemize}

    \item \textbf{States:} Our Environment $E$ is represented by a 3D scan and $s$ is a 3D region of interest that contains $P$. A frame history buffer is used for storing the last planes from previous steps to stabilize search trajectories and prevent getting stuck in repeated cycles. We choose a history size of $4$ frames similar to~\cite{mnih2015human}.

    \item \textbf{Actions:} The agent interacts with $E$ by taking action steps $a \in A$ to modify the position parameters of the plane. The action space consists of eight actions, $\{\pm a_{\theta_x}, \pm a_{\theta_y}, \pm a_{\theta_z}, \pm a_d\}$, which update the plane parameters $a=cos(\theta_x + a_{\theta_x})$, $b=cos(\theta_y + a_{\theta_y})$,  $c=cos(\theta_z + a_{\theta_z})$ and $d = d + a_d$.

    % $(\pm a_{\theta_x}, \pm a_{\theta_y}, \pm a_{\theta_z}, \pm a_d)$, where $a=cos(\theta_x\pm a_{\theta_x})$, $b=cos(\theta_y)$,  $c=cos(\theta_z)$ and $a_d$ controls $d$.

    \item \textbf{Reward:} %Designing an appropriate RL reward function is often difficult.
    The RL reward function forms a proxy for the true task goal and care must be taken to capture exactly what this goal entails. %If an RL algorithm is allowed to overfit to the specified reward, it typically results in undesirable or unexpected results. 
    In our problem instance, the difficulty comes from designing a reward that encourages the agent to move towards the target plane while still being learnable. 
    With these considerations, we define the reward $R = \text{sgn}(D(P_{i-1},P_t)-D(P_i,P_t))$, where $D$ is a function to take the Euclidean distance between plane parameters. We further denote $P_i$ as the current predicted plane at step $i$, with $P_t$ the target ground truth plane. The difference of the parameter distances, between the previous and current steps, signifies whether the agent is moving closer to or further away from the desired plane parameters. $R \in \{+1,0,-1\}$ provide the agent with a per step (non-sparse) reward signal, with zero-valued $R$ presents plane oscillations around the correct solution.

    \item \textbf{Terminal State:} The final state is defined as the state in which the agent finds the target plane $P_{t}$. A trigger action can be used to signal when the target state is reached~\cite{maicas2017deep}. However, adding extra actions increases the action space size, which may in turn increase the complexity of the task to be learned. The maximum number of interactions should also be defined in such a setting. We found that terminating the episode when oscillation is detected heuristically works in practice without the need to expand the action space. However, in contrast to \cite{ghesu2017multi}, we choose the terminating action with the lower $Q$-value. We find that $Q$-values are lower when the target plane is closer. Intuitively, the DQN encourages awarding higher $Q$-values to actions when the current plane is far from the target.
% \todoKainz{the last two parts mix in already some discussion at the item's ends. These parts should perhaps be moved down to Discussion.}
\end{itemize}
% =============================================================================
% \subsection{Multi-scale Agent}
\noindent\textbf{Multi-scale Agent:}
In order to provide more structural information, we introduce a novel multi-resolution approach in a coarse-to-fine fashion with hierarchical action steps. In this scenario, $E$ samples a grid of a fixed plane size $(P_{x},P_{y},P_{z})$ of voxels around the plane origin $P_{o}$ and initial spacing $(S_{x},S_{y},S_{z})$ mm. Initially, the agent searches for the plane with higher action steps. Once the target plane is found, $E$ samples the new planes with smaller spacing and the agent uses smaller action steps. Coarser levels in the hierarchy provide additional guidance to the optimization process by enabling the agent to see larger context of the image. Whereas, finer scales provide sharper adjustments for the final estimation of the plane. Similarly, larger step actions speed up the solution towards the target plane, while smaller steps fine tune the final estimation of plane parameters. The same DQN is shared between all levels in the hierarchy, see Fig.~\ref{fig_framework}. The next section exhibits results of utilizing this multi-scale approach.

% =============================================================================

\section{Experiments and Results}
\label{sec_experiments}
The proposed algorithm is assessed using $12$ different experiments; a combination of four different DQN-based methods with three target planes. We evaluate our results using the distance between anatomical landmarks and the detected planes. We also measure the orientation error by calculating the angle between normal vectors of the detected and target planes. 

\begin{figure}[!htb]
  \centering
%   \vspace{-1cm}
  \subfloat[Axial ACPC plane]{\makebox[3.5cm][c]{\includegraphics[height=2.3cm]{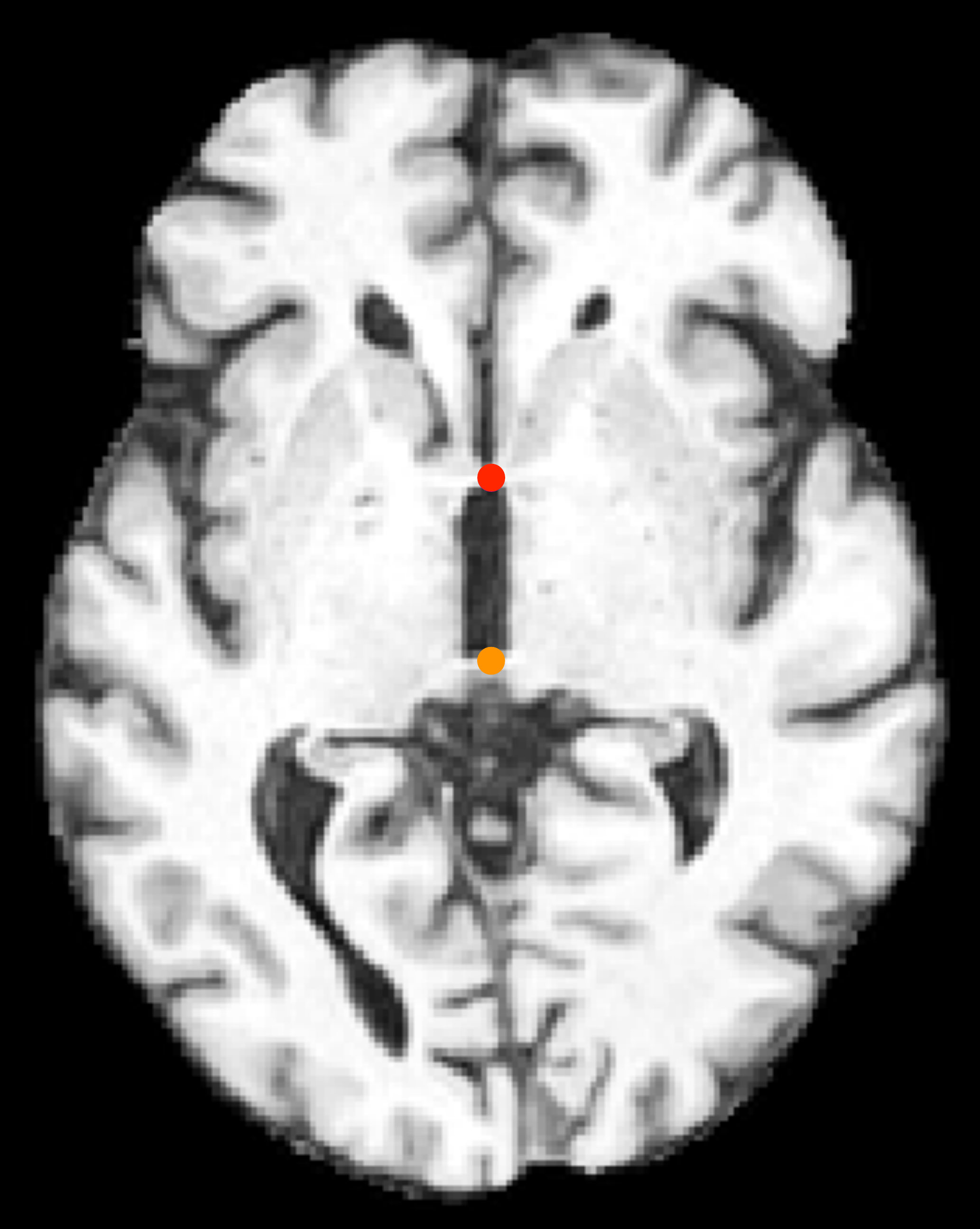}}}\hspace{0.5cm}
  \label{subfig_ACPC}
  \subfloat[Mid-sagittal plane]{\makebox[3.5cm][c]{\includegraphics[height=2.3cm]{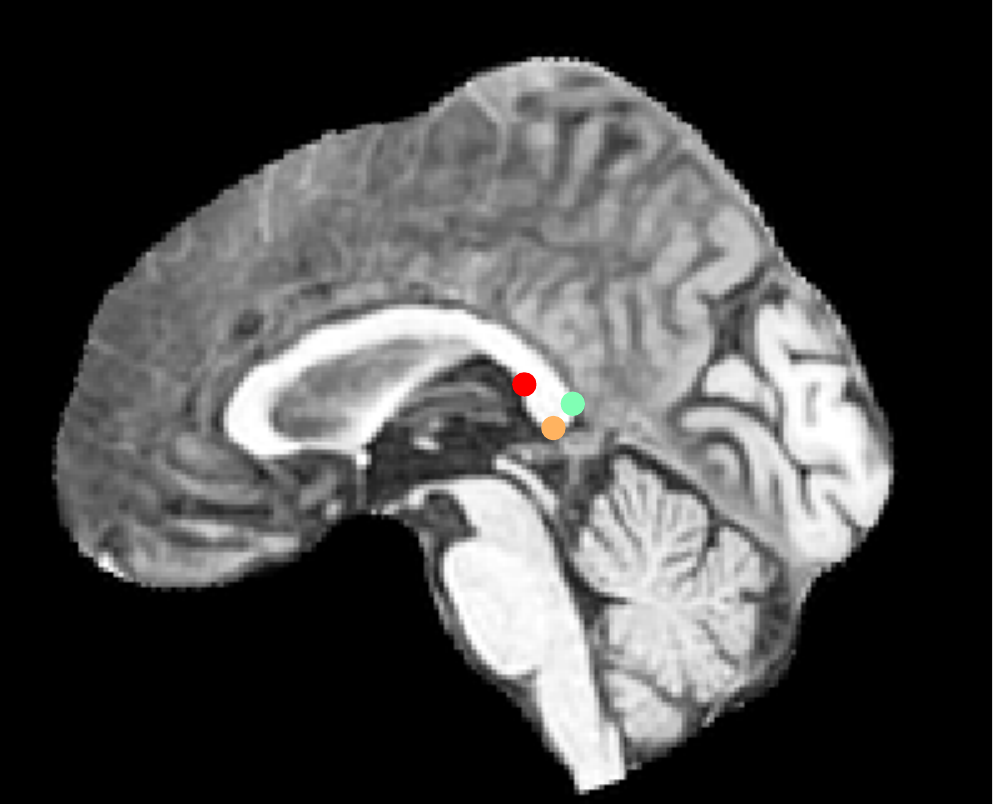}}}\hspace{0.5cm} \label{subfig_midsag}
  \subfloat[4-Chamber plane]{\makebox[3.5cm][c]{\includegraphics[height=2.3cm]{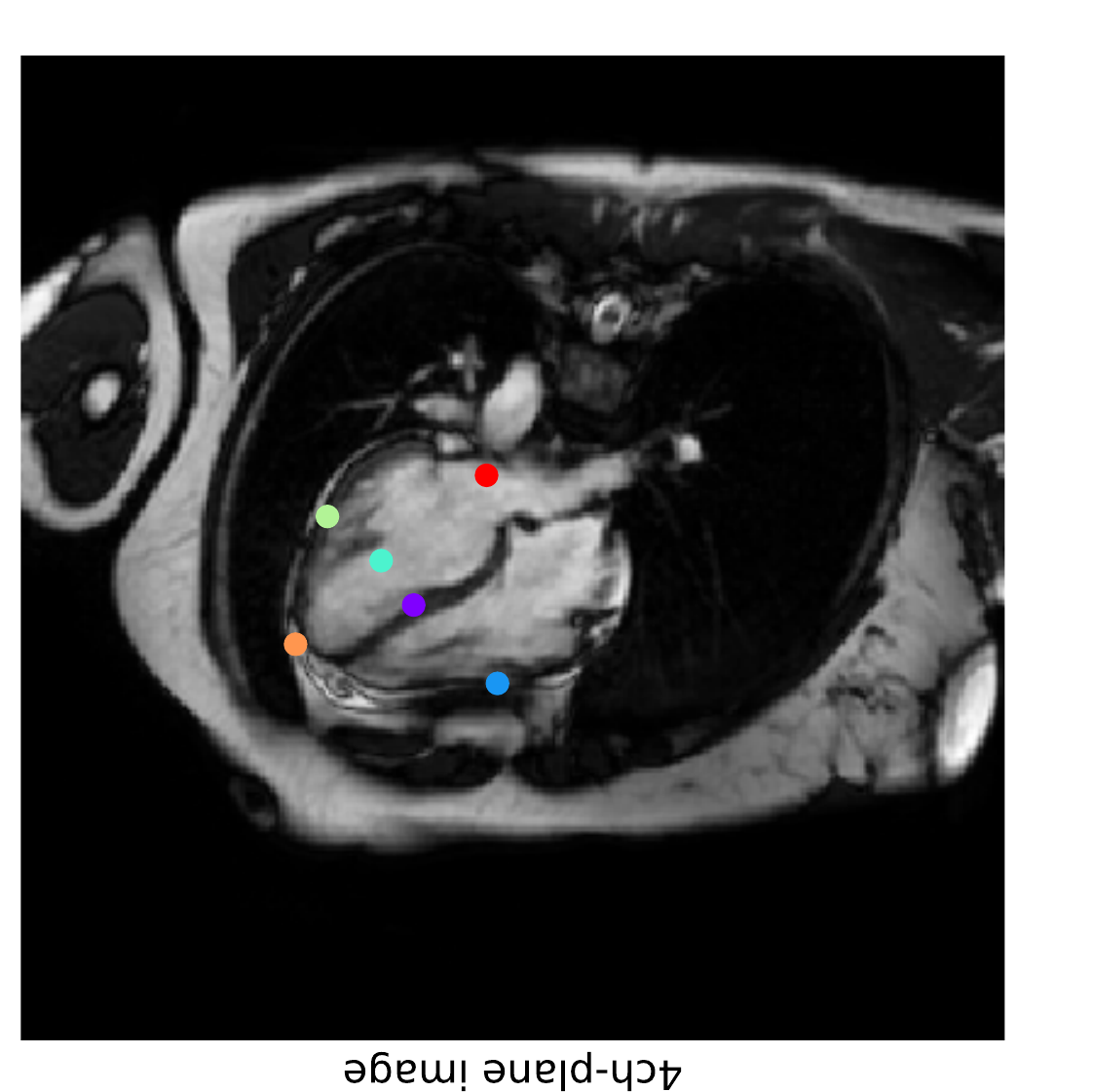}}}\hspace{0.5cm} \label{subfig_4ch}
  
  \caption{Ground truth planes from brain and cardiac MRI scans. (a) ACPC axial plane marking AC (red) and PC (yellow) points. (b) Mid-sagittal plane with outer aspect (green), inferior tip (yellow) and inner aspect (red) points of splenium of corpus callosum. (c) 4-Chamber view with the projected two RV insertion points(violet, green), RV and LV lateral wall turning points (blue, lime), apex (orange), and the center of the mitral valve (red).}
  \label{fig_planes}
%   \vspace{-0.75cm}
\end{figure}

\textbf{Datasets:}
A set of $832$ isotropic $1$mm MR scans were obtained from the ADNI database \cite{mueller2005alzheimer} to evaluate the proposed method. While, a subset of $728$ and $104$ images are used for training and testing. All brain images were skull stripped and affinely aligned to the MNI space, thus allowing ground truth planes to be extracted in the standard directions. 
% ACPC plane will be in the axial direction containing AC and PC points. While the mid-sagittal plane is perpendicular to the ACPC plane in the sagittal direction and passing through Splenium of corpus callosum (outer aspect, inferior tip and inner aspect) points. 
For cardiac images, we use $455$ short-axis cardiac MR of resolution $(1.25 \times 1.25 \times 2)$ mm obtained from the UK Digital Heart Project~\cite{de2014population}. A subset of $364$ and $91$ images are used for training and testing. ACPC planes are evaluated using the AC and PC landmarks for the distance error calculation. Similarly, we use the outer aspect, inferior tip and inner aspect points of splenium of corpus callosum for mid-sagittal planes. For cardiac MRI, we use six landmarks projected on the 4-chamber plane; the two right ventricle (RV) insertion points, right and left ventricles (LV) lateral wall turning points, apex, and the center of the mitral valve, See Fig.~\ref{fig_planes}. %All landmarks, shown in Fig.~\ref{fig_planes}, were manually selected by an expert observer.

\textbf{Experiments:}
During training, a random point is sampled from the 3D input image. The initial random plane is then defined using the normal vector between the center of the image and the random point. The origin of this plane is the projected point of the center of the input image. Finally, a plane of size $(50,50,9)$ voxels is sampled around the plane origin with initial spacing $3 \times 3 \times 3$ mm. Initial $a_{\theta_{x}},a_{\theta_{y}},a_{\theta_{z}}$ equal $8$ and $a_{d}$ equals $4$. With every new scale $a_{\theta_{x}},a_{\theta_{y}},a_{\theta_{z}}$ decrease by a factor of $2$ and $a_{d}$ decrease $1$ unit. $3$-levels of scale with spacing from $3$ to $1$ mm are used for the brain experiments, and $4$-levels of scale from $5$ to $2$ mm for the cardiac experiment. For experiments on cardiac images, initial planes are sampled randomly from the 3D input image within $20\%$ around the center of the image, to avoid sampling outside the field of view.
% =============================================================================

\textbf{Results:}
During inference, the environment samples a plane and the agent updates sequentially new plane's parameters until reaching the terminal state. In order to have a fair comparison between different variants of the proposed method, we fix the initial plane for all models during evaluation. Table~\ref{tab_results} shows the results from these comparative experiments. All methods share similar performance including speed and accuracy, and there is no unique winner for the best overall method. Best performing agents for detecting the mid-sagittal and ACPC planes achieve accuracy of $1.53\pm2.2$ mm and $2.44\pm5.04^{\circ}$, and $1.98\pm2.23$ mm and $4.48\pm14.0^{\circ}$, respectively. Where in cardiac, the task is more complex due to the lower quality and higher variability between different scans. The agent has to navigate in a bigger field of view compared to brain images. Thus Duel DQN-based architectures achieve the best results for detecting the 4-chamber plane with $4.84\pm3.03$ mm and $8.86\pm12.42^{\circ}$ accuracy, as a result from learning a better state value function by decoupling it from action-value function. These results are better than the state-of-the-art~\cite{lu2011automatic}, which achieves an accuracy of $5.7\pm8.5$mm and $17.6\pm19.2^{\circ}$. Unlike~\cite{lu2011automatic}, our method does not require manual annotation of landmarks. More visualization results are published on our \href{https://git.io/vhuMZ}{github}.
\begin{table}[!htb]
% \vspace{-0.6cm}
\centering
\caption{Results of our multi-scale RL agent detecting 3 different MRI planes.}
\label{tab_results}
\resizebox{0.85\textwidth}{!}{
    	\begin{tabular}{ccccccc}

      	\toprule[1pt]
            \textbf{Model}  & \multicolumn{2}{c}{\textbf{Mid-sagittal brain}} & \multicolumn{2}{c}{\textbf{ACPC brain}} 
                            & \multicolumn{2}{c}{\textbf{4-Chamber cardiac}} \\
        
            \cmidrule(lr){1-7}
            & $e_{d}(mm)$ & $e_{\theta}(^{\circ})$ & $e_{d}(mm)$ & $e_{\theta}(^{\circ})$ & $e_{d}(mm)$ & $e_{\theta}(^{\circ})$ \\
    
            \cmidrule(lr){2-3}  \cmidrule(lr){4-5}  \cmidrule(lr){6-7} 
              
            \multicolumn{1}{l}{\textbf{DQN}}        & $1.65\pm1.99$ & $2.42\pm5.27$
                                                    & $2.61\pm5.44$ & \textbf{3.23}$\pm$\textbf{6.03} 
                                                    & $5.61\pm4.09$ & $10.16\pm10.62$ \\
                              
            \multicolumn{1}{l}{\textbf{DDQN}}       & $2.08\pm2.58$ & $3.44\pm7.46$ 
                                                    & \textbf{1.98}$\pm$\textbf{2.23} & $4.48\pm14.00$
                                                    & $5.79\pm4.58$ & $11.20\pm14.86$ \\
                                    
            \multicolumn{1}{l}{\textbf{Duel DQN}}   & $1.69\pm1.98$ & $3.82\pm7.15$ 
                                                    & $2.13\pm1.99$ & $5.24\pm13.75$
                                                    & \textbf{4.84}$\pm$\textbf{3.03} & $8.86\pm12.42$ \\
                              
            \multicolumn{1}{l}{\textbf{Duel DDQN}}  & \textbf{1.53}$\pm$\textbf{2.20} & \textbf{2.44}$\pm$\textbf{5.04} 
                                                    & $5.30\pm11.19$ & $5.25\pm12.64$ 
                                                    & $5.07\pm3.33$ & \textbf{8.72}$\pm$\textbf{7.44} \\
            \bottomrule[1pt]
        \end{tabular}
}
% \vspace{-0.6cm}
\end{table}

% =============================================================================
\textbf{Implementation}
Training times are around $12-24$ hours for the brain experiments and $2-4$ days for the cardiac experiments using an NVIDIA GTX 1080Ti GPU. During inference, the agent finds the target plane using iterative steps, where each step takes {\raise.17ex\hbox{$\scriptstyle\mathtt{\sim}$}}$0.02$s. The details of the our proposed network for DQN are in Figure~\ref{fig_framework}. The source code of our implementation is publicly available on github \href{https://git.io/vhuMZ}{https://git.io/vhuMZ}.
% \subsubsection*{Acknowledgments.}
% =============================================================================
% =============================================================================
\section{Discussion and Conclusion}
We proposed a novel approach based on multi-scale reinforcement learning agents for automatic standard view extraction. Our approach is capable of finding standardized planes in real time, which in turn enables accelerated image acquisition. Consequently, it can alleviate the comparison between different imaging examinations using anatomically standardized biometric measurements. We extensively evaluated several DQN based strategies for the detection of three different planes. Our approach achieved good results for the automatic detection of the ACPC and mid-sagittal planes from brain MRI with distance error less than $2$ mm, and for the detection of the 4-chamber plane from cardiac MRI with distance error around $5$ mm.
% =============================================================================

\textbf{Limitations:} 
Our results show that the optimal algorithm for achieving the best performance is environment-dependant. In general, reinforcement learning is a difficult problem that needs a careful formulation of its elements such as states, rewards and actions. For example, RL tends to overfit to the reward signals, which may cause unexpected behaviours. Therefore the design of the reward function has to capture exactly the desired task, and still be learnable.
% =============================================================================

\textbf{Future Work:} we will investigate using a continuous action space to improve the performance through reduction of quantization errors introduced by fixed action steps. We will also explore the use of either competitive or collaborative multi-agents to detect the same or different anatomical planes. Another future direction is inspired by AlphaGo~\cite{silver2016mastering}, where an RL agent could mimic the moves of a human expert and accumulate this experience, thus learning from experienced operators during real time observation. 
% =============================================================================
\bibliographystyle{splncs03}
\bibliography{references}

\end{document}